\documentclass{article}

 \usepackage[final]{nips_2017}

\usepackage[utf8]{inputenc} 
\usepackage[T1]{fontenc}    
\usepackage{hyperref}       
\usepackage{url}            
\usepackage{booktabs}       
\usepackage{amsfonts}       
\usepackage{nicefrac}       
\usepackage{microtype}      

\usepackage{natbib}

\usepackage{graphicx} 
\usepackage{epsfig}
\usepackage{amssymb}
\usepackage{amsmath}
\usepackage{amsthm}
\usepackage{amsfonts}
\usepackage{bbding}
\usepackage{array}
\usepackage{caption,tabularx,booktabs}
\usepackage{arydshln}
\usepackage{xargs}                      
\usepackage[dvipsnames]{xcolor}  

\usepackage{enumitem}

\usepackage{algorithm}
\usepackage{algorithmic}
\usepackage{float}
\floatstyle{plain} 
\restylefloat{figure}

\usepackage{hyperref}
\usepackage{breakurl}

\newtheorem{thm}{Theorem}
\newtheorem{lem}{Lemma}

\newtheorem{cor}{Corollary}

\newtheorem{rem}{Remark}
\newtheorem{ass}{Assumption}

\makeatletter
\newcounter{subthm} 
\let\savedc@thm\c@hyp

\newcommand{\normhyp}{%
  \let\c@hyp\savedc@hyp 
  \renewcommand\thehyp{\arabic{hyp}}%
} 
\makeatother

\makeatletter
\newcounter{subass} 
\let\savedc@ass\c@hyp

\makeatother

\newcommand\tagthis{\addtocounter{equation}{1}\tag{\theequation}}
\DeclareMathOperator{\Exp}{\mathbb{E}}           
\DeclareMathOperator{\R}{\mathbb{R}} 
\DeclareMathOperator{\Ocal}{\mathcal{O}} 
\newcommand{\eqdef}{\stackrel{\text{def}}{=}}
\newcommand{\setn}{[n]}
\renewcommand{\top}{T}

\newcolumntype{C}[1]{>{\centering\let\newline\\\arraybackslash\hspace{0pt}}m{#1}}

\usepackage{array}
\usepackage{caption,tabularx,booktabs}
\usepackage{arydshln}

\usepackage{xargs}                      
\usepackage[colorinlistoftodos,prependcaption,textsize=tiny]{todonotes}

\newcommandx{\unsure}[2][1=]{\todo[inline,linecolor=red,backgroundcolor=red!25,bordercolor=red,#1]{#2}}
\newcommandx{\change}[2][1=]{\todo[linecolor=blue,backgroundcolor=blue!25,bordercolor=blue,#1]{#2}}
\newcommandx{\info}[2][1=]{\todo[linecolor=OliveGreen,inline,backgroundcolor=OliveGreen!25,bordercolor=OliveGreen,#1]{#2}}
\newcommandx{\improvement}[2][1=]{\todo[linecolor=Plum,inline,backgroundcolor=Plum!25,bordercolor=Plum,#1]{#2}}

\title{Stochastic Recursive Gradient Algorithm for Nonconvex Optimization}

\author{
  Lam M. Nguyen \\
  Industrial and Systems Engineering \\
            Lehigh University, USA \\
  \texttt{lamnguyen.mltd@gmail.com} \\
  \And
  Jie Liu \\
  Industrial and Systems Engineering \\
            Lehigh University, USA \\
  \texttt{jie.liu.2018@gmail.com} \\   
  \And
  Katya Scheinberg \\
  Industrial and Systems Engineering \\
            Lehigh University, USA \\
  On leave at University of Oxford, UK \\          
  \texttt{katyas@lehigh.edu} \\  
  \And
  Martin Tak\'{a}\v{c} \\
  Industrial and Systems Engineering \\
            Lehigh University, USA \\
  \texttt{Takac.MT@gmail.com} \\  
}

\begin{document}

\maketitle

\begin{abstract}
In this paper, we study and analyze the mini-batch version of StochAstic Recursive grAdient algoritHm (SARAH), a method employing the stochastic recursive gradient, for solving empirical loss minimization for the case of nonconvex losses. We provide a sublinear convergence rate (to stationary points) for general nonconvex functions and a linear convergence rate for gradient dominated functions, both of which have some advantages compared to other modern stochastic gradient algorithms for nonconvex losses.

\end{abstract}

\section{Introduction}\label{introduction}

We are interested in the following finite-sum minimization problem 
\begin{gather}\label{eq:problem}
\min_{w\in\R^d} \left\{ P(w) \eqdef \frac{1}{n} \sum_{i\in\setn} f_i(w) \right\},
\end{gather}
where each $f_i$, $i \in \setn\eqdef\{1,\dots,n\}$, is smooth but can be nonconvex, and $P$ is also not necessarily convex. Throughout the paper, we assume that there exists a global optimal solution $w_{*}$ of \eqref{eq:problem}; in other words, there exists a lower bound $P(w_{*})$ of \eqref{eq:problem},  however we do not  assume the knowledge of this bound and we do not seek convergence to $w_*$, in general.

Problems of form~\eqref{eq:problem} cover a wide range of convex and nonconvex problems including but not limited to logistic regression, multi-kernel learning, conditional random fields, neural networks, etc. In many of these applications, the number $n$ of individual components is very large, which makes the exact computation of $P(w)$ and its derivatives  and thus the use of  gradient descent (GD)~\cite{Nocedal2006NO} to solve \eqref{eq:problem} expensive. 

A traditional approach is to employ stochastic gradient descent (SGD)~\cite{RM1951,pegasos}. 
Recently, a large number  of improved variants of  stochastic gradient algorithms have emerged, including SAG/SAGA~\cite{SAGjournal,SAGA}, MISO/FINITO~\cite{MISO,FINITO}, SDCA~\cite{SDCA}, SVRG/S2GD~\cite{SVRG,konecny2015mini}, SARAH~\cite{nguyen2017sarah}~\footnote{Note that numerous modifications of  stochastic gradient algorithms have been proposed, including non-uniform sampling, acceleration, repeated scheme and asynchronous parallelization. In this paper, we refrain from checking and analyzing those variants, and compare only the primary methods.}. While, nonconvex problems of the 
form \eqref{eq:problem} are now  widely used due to the recent interest in deep neural networks, the majority of 
methods are designed and analyzed for  the convex/strongly convex cases. Limited results have been developed for the nonconvex problems~\cite{nonconvexSVRG,nonconvexSVRGZhu,Natasha}, in particular, \cite{nonconvexSVRG,nonconvexSVRGZhu} introduce nonconvex SVRG, and Natasha~\cite{Natasha} is a new algorithm but a variant of SVRG for nonconvex optimization.

In this paper we develop  convergence rate analysis  of a mini-batch variant SARAH for nonconvex problems of the form \eqref{eq:problem}. SARAH has been introduced in~\cite{nguyen2017sarah} and shown to have a sublinear rate of convergence for general convex functions, and a linear rate of convergence for strongly convex functions. As the SVRG method, SARAH has an inner and an outer loop. It has been shown in  ~\cite{nguyen2017sarah}  that, unlike the inner loop of SVRG, the inner loop of SARAH converges. Here we explore the properties of the inner loop of SARAH for general nonconvex functions and show that it converges at the same rate as SGD, but under weaker assumptions and with better constants in the convergence rate. We then analyze  the full SARAH algorithm 
in  the case of gradient dominated functions as a special class of nonconvex functions~\cite{polyak1963gradient,nesterov2006cubic,nonconvexSVRG} for which we show linear convergence to a global minimum.  We will provide the definition of a gradient dominated function in Section \ref{sec_conv_analysis}. We also note that this type of function includes the case where the objective function $P$ is strongly convex, but the component functions $f_i$, $i \in [n]$, are not necessarily convex. 

We now summarize the complexity results of SARAH and other existing methods  for nonconvex functions  in Table \ref{table:summary}. All complexity estimates are in terms of the number of calls to the {\em incremental first order oracle} (IFO) defined in \cite{icml2015_agarwal15}, in other words computations of $(f_i(w), \nabla f_i(w))$ for some $i\in \setn$. 
The iteration complexity analysis aims to bound the number of iterations $\mathcal{T}$, which is needed to guarantee that 
$\|\nabla P(w_\mathcal{T})\|^2\leq \epsilon$. In this case we will say that $w_\mathcal{T}$ is an $\epsilon$-accurate solution. 
However, it is  common practice for  
stochastic gradient
 algorithms to obtain the bound on the number of IFOs after which the algorithm can be terminated with the guaranteed the bound on the  expectation, as follows,
 \begin{equation}\label{eq:accuracy}
\Exp [\| \nabla P(w_\mathcal{T}) \|^2] \leq \epsilon.
\end{equation}
It is important to note that for the stochastic algorithms discussed here, the output  $ w_\mathcal{T}$ is not the last iterate computed by the algorithm, but a randomly selected iterate from the computed sequence. 

Let us discuss the results  in Table \ref{table:summary}. The analysis of SGD in \cite{Ghadimi2013} in performed under  the assumption that 
 $\| \nabla f_i(\cdot) \|\leq \sigma$, for all $i \in \setn$, for some fixed constant $\sigma$. This limits the applicability of the convergence results for SGD and adds dependence on $\sigma$ which can be large.  In contrast, convergence rate of SVRG only requires $L$-Lipschitz continuity of the gradient as does the analysis of SARAH.  
Convergence of SVRG for general nonconvex functions is better than that of the inner loop of SARAH in terms of its dependence on $\epsilon$, but it is worse in term of its dependence on $n$. In addition the bound for SVRG includes an unknown universal constant $\nu$, whose magnitude is not clear and can be quite small. Convergence rate of the full SARAH algorithm for general nonconvex functions remains an open question. 
In the case of $\tau$-gradient dominated functions, full SARAH convergence rate dominates that of the other algorithms.

\begin{table}[h]
 \scriptsize
 \centering 
\caption{Comparisons between different algorithms for nonconvex functions.}
\label{table:summary}
\begin{tabular}{|C{1.15cm}|c|c|c|c| }
\hline 
Method & GD  & SGD  & SVRG & {\textbf{SARAH}} \\
\hline
Nonconvex & $\Ocal\left(\tfrac{nL}{\epsilon} \right)$ & $\Ocal\left(\tfrac{L \sigma^2}{\epsilon^2}\right)$ & $\Ocal\left(n + \tfrac{n^{2/3}L}{\nu \epsilon} \right)$ & $\Ocal\left(n + \tfrac{L^2}{\epsilon^2} \right)$\\
\hline
$\tau$-Gradient Dominated & $\Ocal\left(nL \tau \log(\tfrac{1}{\epsilon}) \right)$ & $\Ocal\left(\tfrac{L \tau \sigma^2}{\epsilon^2}\right)$ & $\Ocal\left((n + \frac{n^{2/3} L \tau}{\nu} ) \log(\tfrac{1}{\epsilon}) \right)$  & $\Ocal\left((n + L^2\tau^2  ) \log(\tfrac{1}{\epsilon}) \right)$\\
\hline
\end{tabular}
\end{table}

(GD (\cite{nesterov2004,nonconvexSVRG}), SGD (\cite{Ghadimi2013,nonconvexSVRG}), SVRG (\cite{nonconvexSVRG}))

\textbf{Our contributions.} In summary, in this paper we analyze SARAH with mini-batches for nonconvex optimization. SARAH originates from the idea of momentum SGD, SAG/SAGA, SVRG and L-BFGS and is initially proposed for convex optimization, and is now proven to be effective for minimizing finite-sum problems of general nonconvex functions. We summarize the key contributions of the paper as follows.
\begin{itemize}[noitemsep,nolistsep]
\item We study and extend SARAH framework~\cite{nguyen2017sarah} with \emph{mini-batches} to solving \emph{nonconvex} loss functions, which cover the popular deep neural network problems. We are able to provide a sublinear convergence rate of the inner loop of SARAH for general nonconvex functions, under milder assumptions than that of SGD. 

\item Like SVRG~\cite{nonconvexSVRG}, SARAH algorithm is shown to enjoy  linear convergence rate for $\tau$-gradient dominated functions--a special class of possibly nonconvex functions~\cite{polyak1963gradient,nesterov2006cubic}. 
\item Similarly to SVRG, SARAH maintains a constant learning rate for nonconvex optimization, and a larger mini-batch size allows the use of a more aggressive learning rate and a smaller inner loop size.
\item Finally, we present numerical results, where a practical version of SARAH, introduced in~\cite{nguyen2017sarah} is shown to be competitive on standard  neural network training tasks. 
\end{itemize}

\section{Stochastic Recursive Gradient Algorithm}

The pivotal idea of SARAH, like many existing algorithms, such as SAG, SAGA and BFGS~\cite{Nocedal2006NO},  is to utilize past  stochastic gradient estimates to improve convergence. In contrast with SAG, SAGA and BFGS~\cite{Nocedal2006NO}, SARAH does not store  past information thus significantly reducing  storage cost. 
We present SARAH as a two-loop  algorithm in Figure~\ref{sarah}, with  SARAH-IN  in Figure~\ref{sarah_1} describing the inner  loop. 


\begin{figure}[H]
  \centering
  \caption{Algorithm SARAH}
   \label{sarah}
  \fbox{\begin{minipage}{0.80\columnwidth}
\begin{algorithmic}
   \STATE {\bfseries Input:} $\tilde{w}_0$, the learning rate $\eta > 0$, the batch size $b$ and the inner loop size $m$.
   \STATE {\bfseries Iterate:}
   \FOR{$s=1,2,\dots$}
   \STATE $\tilde{w}_s = \text{SARAH-IN}(\tilde{w}_{s-1},\eta,b,m)$
   \ENDFOR
   \STATE \textbf{Output:} $\tilde{w}_s$
\end{algorithmic}
\end{minipage}}
\end{figure}

\begin{figure}[H]
  \centering
  \caption{Algorithm SARAH within a single outer loop: SARAH-IN($w_0,\eta,b,m$)}
   \label{sarah_1}
  \fbox{\begin{minipage}{0.92\columnwidth}
\begin{algorithmic}
   \STATE {\bfseries Input:} $w_0 (= \tilde{w}_{s-1})$, the learning rate $\eta > 0$, the batch size $b$ and the inner loop size $m$.
   \STATE Evaluate the full gradient: $v_0 = \frac{1}{n}\sum_{i=1}^{n} \nabla f_i(w_0)$
   \STATE Take a gradient descent step: $w_1 = w_0 - \eta v_0$
   \STATE {\bfseries Iterate:}
   \FOR{$t=1,\dots,m-1$}
   \STATE Choose  a mini-batch $I_{t} \subseteq \setn$ of size $b$ uniformly at random (without replacement)
   \STATE Update the stochastic recursive gradient: \begin{equation}\label{eq:vt_mb}
   v_{t} = \frac{1}{b} \sum_{i \in I_{t}} [\nabla f_{i} (w_{t}) - \nabla f_{i}(w_{t-1})] + v_{t-1}
   \end{equation}
   \STATE Update the iterate: $w_{t+1} = w_{t} - \eta v_{t}$
   \ENDFOR
   \STATE $\tilde{w} = w_{t}$ with $t$ chosen uniformly randomly from $\{0,1,\dots,m\}$
   \STATE \textbf{Output:} $\tilde{w}$
\end{algorithmic}
\end{minipage}}
\end{figure}

Similarly to SVRG, in each outer iteration, SARAH proceeds with the evaluation of a full gradient followed by an inner loop of $m$ stochastic steps.  
SARAH requires one computation of the full gradient at the start of its inner loop and then proceeds by updating this gradient information using stochastic gradient estimates over $m$ inner steps. Hence, each outer iteration corresponds to a cost of $\mathcal{O}(n+bm)$ component gradient evaluations (or IFOs).   For simplicity let us consider the inner loop update for $b=1$, as presented in~\cite{nguyen2017sarah}:
\begin{equation}\label{eq:vt}
   v_{t} = \nabla f_{i_{t}} (w_{t}) - \nabla f_{i_{t}}(w_{t-1}) + v_{t-1},
\end{equation}
Note that unlike SVRG, which uses the gradient updates $ v_{t} = \nabla f_{i_{t}} (w_{t}) - \nabla f_{i_{t}}(w_{0}) + v_{0}$, SARAH's gradient estimate $v_{t}$
 iteratively includes all past stochastic gradients, however,  SARAH consumes a memory of $\Ocal(d)$ instead of $\Ocal(nd)$ in the cases of SAG/SAGA and BFGS,
 because this past information is simply averaged, instead of being stored. 

%
%
%
%

With either $m=1$ or $s=1$ and $b = n$, the algorithm SARAH recovers gradient descent (GD). We remark here that we also recover the convergence rate theoretically for GD with $s=1$ and $b=n$ 
In the following section, we analyze theoretical convergence properties of SARAH when applied to nonconvex functions.

\section{Convergence Analysis}\label{sec_conv_analysis}

 First, we will introduce the sublinear convergence of SARAH-IN for general nonconvex functions. Then we present the linear convergence of SARAH over a special class of gradient dominated functions \cite{polyak1963gradient,nesterov2006cubic,nonconvexSVRG}. Before proceeding to the analysis, let us start by stating some assumptions.

\begin{ass}[$L$-smooth]
\label{ass_Lsmooth}
Each $f_i: \mathbb{R}^d \to \mathbb{R}$, $i \in \setn$, is $L$-smooth, i.e., there exists a constant $L > 0$ such that
\begin{gather*}
\| \nabla f_i(w) - \nabla f_i(w') \| \leq L \| w - w' \|, \ \forall w,w' \in \mathbb{R}^d. \tagthis\label{eq:Lsmooth_basic}
\end{gather*} 
\end{ass}

Assumption \ref{ass_Lsmooth} implies that $P$ is also $L$-smooth. Then, by the property of $L$-smooth function (in \cite{nesterov2004}), we have 
\begin{align*}
P(w) &\leq P(w') + \nabla P(w')^T(w-w') + \frac{L}{2}\|w-w'\|^2, \ \forall w, w' \in \mathbb{R}^d.  
\tagthis\label{eq:Lsmooth}
\end{align*}

The following assumption will be made only when  appropriate, otherwise, it will be dropped. 

\begin{ass}[$\tau$-gradient dominated]
\label{ass_gradientdominated}
$P$ is $\tau$-gradient dominated, i.e., there exists a constant $\tau > 0$ such that $\forall w \in \mathbb{R}^d$, 
\begin{gather*}
P(w) - P(w_{*}) \leq \tau \| \nabla P(w) \|^2, \tagthis \label{gradientdominated}
\end{gather*}
where $w_{*}$ is a global minimizer of $P$. 
\end{ass}

We can observe that every stationary point of the $\tau$-gradient dominated function $P$ is a global minimizer. However, such a function $P$ needs not necessarily be convex. If $P$ is $\mu$-\emph{strongly convex} (but each $f_i$, $i \in \setn$, is possibly nonconvex), then $2\mu [P(w) - P(w_{*})] \leq \| \nabla P(w) \|^2$, $\forall w \in \mathbb{R}^d$. Thus, a $\mu$-strongly convex function is also $1/(2\mu)$-gradient dominated. 


The following two results - Lemmas \ref{lem_main_derivation_mb} and \ref{lem:var_diff_mb} - are essentially the same as Lemmas 1 and 2 in  \cite{nguyen2017sarah}
 with a slight modification to include the case when $b$ is not necessarily equal to $1$. We present the proofs in the supplementary material for completeness. 


\begin{lem}\label{lem_main_derivation_mb}
Suppose that Assumption \ref{ass_Lsmooth} holds. Consider SARAH-IN (SARAH within a single outer loop in Figure~\ref{sarah_1}), then we have 
\begin{align*}
\sum_{t=0}^{m} \mathbb{E}[ \| \nabla P(w_{t})\|^2 ]  \leq \frac{2}{\eta} [ P(w_{0}) - P(w_{*})] + \sum_{t=0}^{m} \mathbb{E}[ \| \nabla P(w_{t}) - v_{t} \|^2 ]  
 - ( 1 - L\eta ) \sum_{t=0}^{m} \mathbb{E} [ \| v_{t} \|^2 ],  \tagthis \label{eq:001}
\end{align*}
where $w_{*}$ is a global minimizer of $P$. 
\end{lem}

\begin{lem}\label{lem:var_diff_mb}
Suppose that Assumption \ref{ass_Lsmooth} holds. Consider $v_{t}$ defined by \eqref{eq:vt_mb} in SARAH-IN, then for any $t\geq 1$, 
\begin{align*}
\mathbb{E}[ \| \nabla P(w_{t}) - v_{t} \|^2 ] 
= \sum_{j = 1}^{t} \mathbb{E}[ \| v_{j} - v_{j-1} \|^2 ]  
 - \sum_{j = 1}^{t} \mathbb{E}[ \| \nabla P(w_{j}) - \nabla P(w_{j-1}) \|^2 ]. 
\end{align*}
\end{lem}

With the above Lemmas, we can derive the following upper bound for $\mathbb{E}[ \| \nabla P(w_{t}) - v_{t} \|^2 ]$. 

\begin{lem}\label{lem:var_diff_mb_02}
Suppose that Assumption \ref{ass_Lsmooth} holds. Consider $v_{t}$ defined by \eqref{eq:vt_mb} in SARAH-IN. Then for any $t\geq 1$, 
\begin{align*}
\mathbb{E}[ \| \nabla P(w_{t}) - v_{t} \|^2 ]  \leq \frac{1}{b} \left( \frac{n-b}{n-1} \right) L^2 \eta^2 \sum_{j=1}^{t} \mathbb{E}[\| v_{j-1} \|^2]. 
\end{align*}
\end{lem}

The proof of Lemma \ref{lem:var_diff_mb_02} is provided in the supplementary material. Using the above lemmas, we are now able to obtain the following convergence rate result for SARAH-IN. 

%

\begin{thm}\label{thm:nonconvex_01_mb}
Suppose that Assumption \ref{ass_Lsmooth} holds. Consider SARAH-IN (SARAH within a single outer loop in Figure~\ref{sarah_1}) with 
\begin{align*}
\eta \leq \frac{2}{L\left(\sqrt{1 + \frac{4m}{b}\left(\frac{n-b}{n-1} \right)} + 1\right)}. \tagthis \label{eta_mb}
\end{align*}

Then we have
\begin{align*}
\mathbb{E}[ \| \nabla P(\tilde{w})\|^2 ] \leq \frac{2}{\eta(m+1)} [ P(w_{0}) - P(w_{*})], 
\end{align*}
where $w_{*}$ is a global minimizer of $P$, and $\tilde{w} = w_t$, where $t$ is chosen uniformly at random from $\{0,1,\dots,m\}$. 
\end{thm}

This result shows a sublinear convergence rate for SARAH-IN  with increasing $m$. Consequently, with $b=1$ and $\eta = \frac{2}{L(\sqrt{1 + 4m} + 1)}$, to obtain 
$$\mathbb{E}[ \| \nabla P(\tilde{w})\|^2 ] \leq \frac{L(\sqrt{1 + 4m} + 1)}{(m+1)} [ P(w_{0}) - P(w_{*})] \leq \epsilon,$$
 it is sufficient to make $m = \Ocal(L^2/\epsilon^2)$. Hence, the total complexity to achieve an $\epsilon$-accurate solution is $(n + 2m) = \Ocal(n+L^2/\epsilon^2)$. Therefore, we have the following conclusion for complexity bound.

\begin{cor}\label{cor:nonconvex_01}
Suppose that Assumption \ref{ass_Lsmooth} holds. Consider SARAH within a single outer iteration with batch size $b = 1$ and the learning rate $\eta = \Ocal(1/(L\sqrt{m}))$ where $m$ is the total number of iterations, then  $\|\nabla P(w_t)\|^2$ converges sublinearly in expectation with a rate of $\Ocal(L/\sqrt{m})$, and therefore, the total complexity to achieve an $\epsilon$-accurate solution defined in \eqref{eq:accuracy} is $\Ocal(n+L^2/\epsilon^2)$.   
\end{cor}

Finally,  we present the result for SARAH with multiple outer iterations in application to  the class of gradient dominated functions defined in \eqref{gradientdominated}. 

%
%

\begin{thm}\label{thm:nonconvex_02_mb}
Suppose that Assumptions \ref{ass_Lsmooth} and \ref{ass_gradientdominated} hold. Consider SARAH (in Figure~\ref{sarah}) with $\eta$ and $m$ such that
\begin{align*}
\eta \leq \frac{2}{L\left(\sqrt{1 + \frac{4m}{b}\left(\frac{n-b}{n-1} \right)} + 1\right)} \ \ \ \ \ \text{and} \ \ \ \ \ \frac{\eta(m+1)}{2}  > \tau . 
\end{align*}

Then we have
\begin{align*}
\mathbb{E}[ \| \nabla P(\tilde{w}_s)\|^2 ] \leq (\bar{\gamma}_m)^s \| \nabla P(\tilde{w}_0)\|^2,
\end{align*}

where
\begin{align*}
\bar{\gamma}_m = \frac{2\tau}{\eta(m+1)} < 1. 
\end{align*}
\end{thm}

Consider the case when $b=1$ and $\eta = \frac{2}{L(\sqrt{1 + 4m} + 1)}$. We need $m = \Ocal\left(L^2\tau^2 \right)$ to satisfy $\frac{\eta(m+1)}{2} = \frac{m + 1}{L\sqrt{1 + 4m} + 1} > \tau$. To obtain $$\mathbb{E}[ \| \nabla P(\tilde{w}_s)\|^2 ] \leq (\bar{\gamma}_m)^s \| \nabla P(\tilde{w}_0)\|^2 \leq \epsilon,$$ it is sufficient to have $s = \Ocal\left(\log(1/\epsilon)\right)$. This implies the total complexity to achieve an $\epsilon$-accurate solution is $(n + 2m)s = \Ocal\left((n+ L^2\tau^2)\log(1/\epsilon)\right)$ and we can summarize the conclution as follows.

\begin{cor}\label{cor:nonconvex_02}
Suppose that Assumptions \ref{ass_Lsmooth} and \ref{ass_gradientdominated} hold. Consider SARAH with parameters from Theorem \ref{thm:nonconvex_02_mb} with batch size $b = 1$ and the learning rate $\eta = \Ocal\left(1/(L\sqrt{m})\right)$, then the total complexity to achieve an $\epsilon$-accurate solution defined in \eqref{eq:accuracy} is $\Ocal\left((n+ L^2\tau^2)\log(1/\epsilon)\right)$.   
\end{cor}

\section{Discussions on the mini-batches sizes}\label{sec_minibatch_sarah}

Let us discuss two simple corollaries of Theorem \ref{thm:nonconvex_01_mb}.  

The first corollary is obtained trivially by substituting the learning rate into the complexity bound in Theorem \ref{thm:nonconvex_01_mb}. 

\begin{cor}\label{cor:nonconvex_01_mb}
Suppose that Assumption \ref{ass_Lsmooth} holds. Consider SARAH-IN (SARAH within a single outer loop in Figure~\ref{sarah_1}) with 
\begin{align*}
\eta = \frac{2}{L\left(\sqrt{1 + \frac{4m}{b}\left(\frac{n-b}{n-1} \right)} + 1\right)}. \tagthis \label{eta_mb_cor}
\end{align*}

Then we have
\begin{align*}
\mathbb{E}[ \| \nabla P(\tilde{w})\|^2 ] \leq \frac{L\left(\sqrt{1 + \frac{4m}{b}\left(\frac{n-b}{n-1} \right)} + 1\right)}{(m+1)} [ P(w_{0}) - P(w_{*})], 
\end{align*}
where $w_{*}$ is a global minimizer of $P$, and $\tilde{w} = w_t$, where $t$ is chosen uniformly at random from $\{0,1,\dots,m\}$. 
\end{cor}

\begin{rem}
We can clearly observe that the rate of convergence for SARAH-IN depends on the size of $b$. For a larger value of $b$, we can use a more aggressive learning rate and it requires the smaller number of iterations to achieve an $\epsilon$-accurate solution. In particular, when $b=n$, SARAH-IN reduces to the GD method and its convergence rate becomes that of gradient descent, 
\begin{align*}
\mathbb{E}[ \| \nabla P(\tilde{w})\|^2 ] \leq \frac{2L}{(m+1)} [ P(w_{0}) - P(w_{*})], 
\end{align*}

and the total complexity to achieve an $\epsilon$-accurate solution is $n \cdot m = \Ocal\left(\tfrac{nL}{\epsilon} \right)$. 
However, the total work in terms of IFOs increases with $b$. When $b \neq n$, the total complexity to achieve an $\epsilon$-accurate solution is $(n + 2bm) = \Ocal\left(n + \tfrac{L^2}{\epsilon^2}  \left(\frac{n-b}{n-1} \right) \right)$. 
\end{rem}

Let us set $m = n-1$ in Corollary~\ref{cor:nonconvex_01_mb}, we can achieve the following result. 

\begin{cor}\label{cor:nonconvex_01_mb_ex00}
Suppose that Assumption \ref{ass_Lsmooth} holds. Consider SARAH-IN with $m = n-1$, and 
$$\eta = \frac{2}{L\left(\sqrt{4(n/b) - 3} + 1\right)}.$$ 

Then we have
\begin{align*}
\mathbb{E}[ \| \nabla P(\tilde{w})\|^2 ] \leq \frac{L\left(\sqrt{4(n/b) - 3} + 1\right)}{n} [ P(w_{0}) - P(w_{*})], 
\end{align*}
where $w_{*}$ is a global minimizer of $P$, and $\tilde{w} = w_t$, where $t$ is chosen uniformly at random from $\{0,1,\dots,n-1\}$. 
\end{cor}

\begin{rem}
For SARAH-IN with the number of iterations $m = n - 1$ and the learning rate $\eta = \Ocal\left(1/(L\sqrt{(n/b)}) \right)$, we could achieve a convergence rate of $\Ocal(L/\sqrt{bn})$. We can observe that the value of $b$ significantly affects the rate. For example, when $b = n/\beta$, $\beta > 1$ and $b = n^{\alpha}$, $\alpha < 1$, the convergence rates become $\Ocal(L\sqrt{\beta}/n)$ and $\Ocal(L/\sqrt{n^{\alpha+1}})$, respectively. 
\end{rem}

\section{Numerical Experiments}\label{sec:exp}

We now turn to the numerical study and conduct experiments on the multiclass classification problem with neural networks, which is the typical challegeing nonconvex problem in machine learning.

\paragraph{SARAH+ as a Practical Variant} \cite{nguyen2017sarah} proposes SARAH+ as a practical variant of SARAH. Now we propose SARAH+ for the nonconvex optimization by running Algorithm SARAH (Figure~\ref{sarah}) with the following SARAH-IN algorithm (Figure~\ref{sarahplus}). Notice that SARAH+ is different from SARAH in that  the inner loop is terminated adaptively  instead of using a fixed choice of the inner loop size $m$. This is idea is based on the fact
 that the norm $\|v_t\|$ converges to zero expectation, which has been both proven theoretically and verified numerically for convex optimization in \cite{nguyen2017sarah}. Under the assumption that similar behavior happens in the nonconvex case, instead of  tuning the inner loop size for SARAH, 
 we believe that  a proper choice of the ratio $\gamma$ below, the  automatic loop termination  can give  superior or competitive performances.

\begin{figure}[H]
  \centering
  \caption{Algorithm SARAH within a single outer loop: SARAH-IN($w_0,\eta,b,m$)}
   \label{sarahplus}
  \fbox{\begin{minipage}{1.0\columnwidth}
\begin{algorithmic}
   \STATE {\bfseries Input:} $w_0 (= \tilde{w}_{s-1})$, the learning rate $\eta > 0$, the batch size $b$ and the maximum inner loop size $m$.
   \STATE Evaluate the full gradient: $v_0 = \frac{1}{n}\sum_{i=1}^{n} \nabla f_i(w_0)$
   \STATE Take a gradient descent step: $w_1 = w_0 - \eta v_0$
   \STATE {\bfseries Iterate:}
   \WHILE{$\|v_{t-1}\|^2 >  \gamma \|v_{0}\|^2$ {\bf and} $t<m$}
   \STATE Choose  a mini-batch $I_{t} \subseteq \setn$ of size $b$ uniformly at random (without replacement)
   \STATE Update the stochastic recursive gradient:
   $$v_{t} = \frac{1}{b} \sum_{i \in I_{t}} [\nabla f_{i} (w_{t}) - \nabla f_{i}(w_{t-1})] + v_{t-1}$$
   \STATE Update the iterate and index: $w_{t+1} = w_{t} - \eta v_{t}$; $t = t + 1$
   \ENDWHILE
   \STATE $\tilde{w} = w_{t}$ with $t$ chosen uniformly randomly from $\{0,1,\dots,m\}$
   \STATE \textbf{Output:} $\tilde{w}$
\end{algorithmic}
\end{minipage}}
\end{figure}

\paragraph{Networks and Datasets} We perform numerical experiments with neural nets with one fully connected hidden layer of $n_h$ nodes, followed by a fully connected output layer which feeds into the softmax regression and cross entropy objective, with the weight decay regularizer ($\ell_2$-regularizer) with parameter $\lambda$. We test the performance on the datasets MNIST~\cite{Lecun98}~\footnote{Available at \burl{http://yann.lecun.com/exdb/mnist/}.} and CIFAR10~\cite{Krizhevsky09}~\footnote{Available at \burl{https://www.cs.toronto.edu/~kriz/cifar.html}.} with $n_h=300, \lambda=$1e-04 and $n_h=100, \lambda=$1e-03, respectively. Both datasets have 10 classes, i.e. 10 softmax output nodes in the network, and are normalized to interval $[0,1]$ as a simple data pre-processing. This network of MNIST achieves the best performance for neural nets with a single hidden layer. Information on both datasets is also available in Table~\ref{table:stats}.


\paragraph{Optimization Details} We compare the efficiency of SARAH, SARAH+~\cite{nguyen2017sarah}, SVRG~\cite{nonconvexSVRG}, AdaGrad~\cite{AdaGrad} and SGD-M (momentum SGD~\cite{Polyak1964,SGD-M})~\footnote{While SARAH, SVRG, SGD have been proven effective for nonconvex optimization, as far as we know, the SGD variants AdaGrad and SGD-M do not have theoretical convergence for nonconvex optimization.} numerically in terms of number of effective data passes, where the last two algorithms are efficient SGD variants  available in the Google open-source library \emph{Tensorflow}~\footnote{See \burl{https://www.tensorflow.org}.}. As the choice of initialization for the weight parameters is very important, we apply a widely used mechanism called \emph{normalized initialization}~\cite{Initialization} where the weight parameters between layers $j$ and $j+1$ are sampled uniformly from $\left[-\sqrt{6/(n_j+n_{j+1})},\sqrt{6/(n_j+n_{j+1})}\right].$ In addition, we use  mini-batch size $b=10$  in all the algorithms.

 \begin{table}[H]
\scriptsize
\centering
\caption{Summary of statistics and best parameters of all the algorithms for the two datasets.}
\label{table:stats}
\begin{tabular}{|c||C{2cm}|C{1.2cm}||C{1.15cm}|C{1.15cm}|C{1.15cm}|C{1.15cm}|C{1.15cm}|}
\hline
Dataset  & Number of Samples $(n_{\text{train}}, n_{\text{test}})$ & Dimensions ($d$) & SARAH ($m^*,\eta^*$) & SARAH+ ($\eta^*$) & SVRG $(m^*,\eta^*)$ & AdaGrad ($\delta^*, \eta^*$)  & SGD-M ($\gamma^*, \eta^*$)   \\
\hline \hline 
\emph{MNIST} & (60,000, 10,000) & 784 & (0.1n, 0.08)& 0.2 & (0.4n, 0.08) & (0.01, 0.1) & (0.7, 0.01)\\ 
\hline 
\emph{CIFAR10} & (50,000, 10,000) & 3072 & (0.4n, 0.03)  & 0.02 & (0.8n, 0.02) & (0.05, 1.0) & (0.7, 0.001)\\
\hline 
\end{tabular}
\end{table}

\begin{figure}[t] 
 \epsfig{file=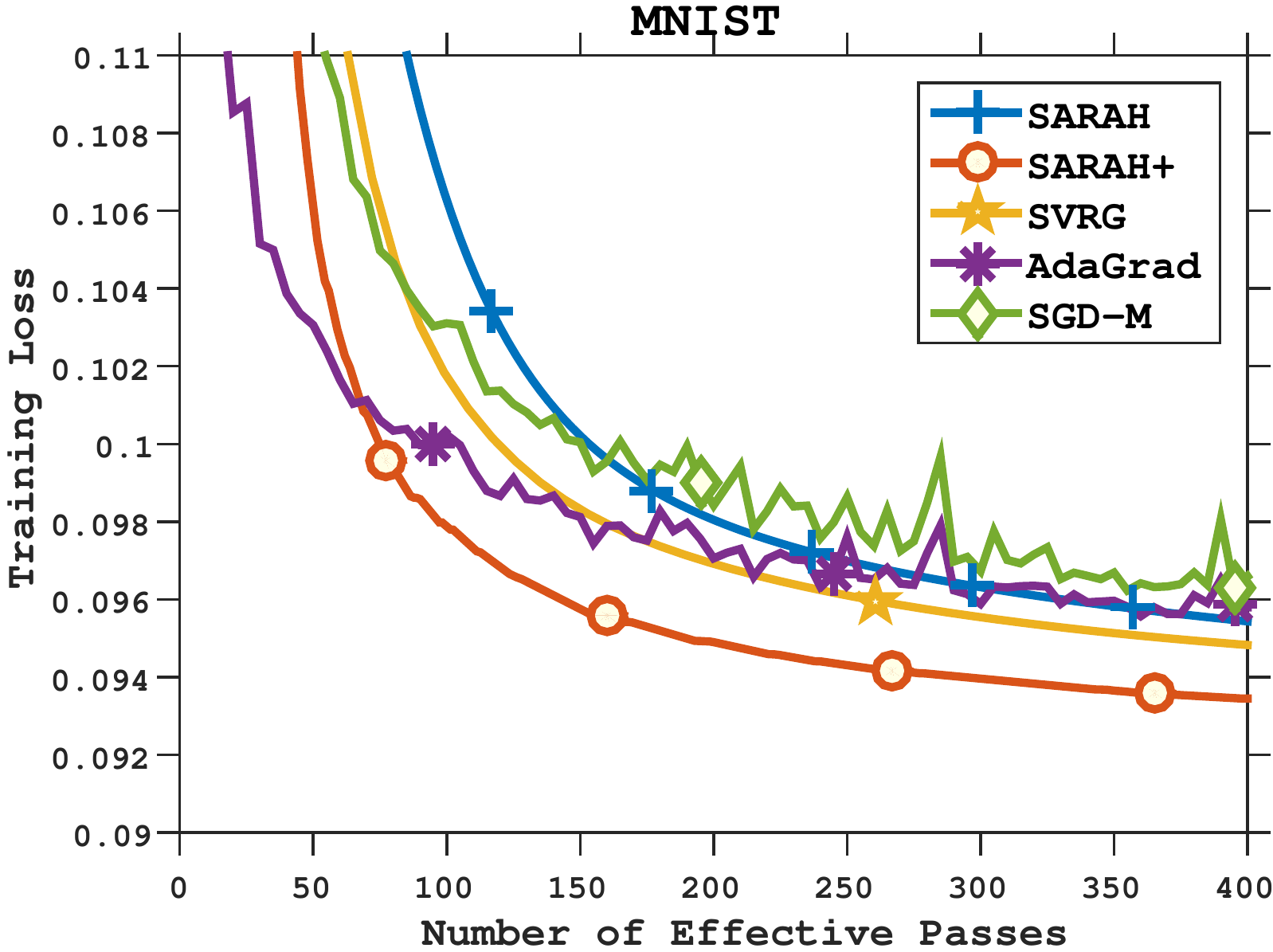,width=0.5\textwidth}
 \epsfig{file=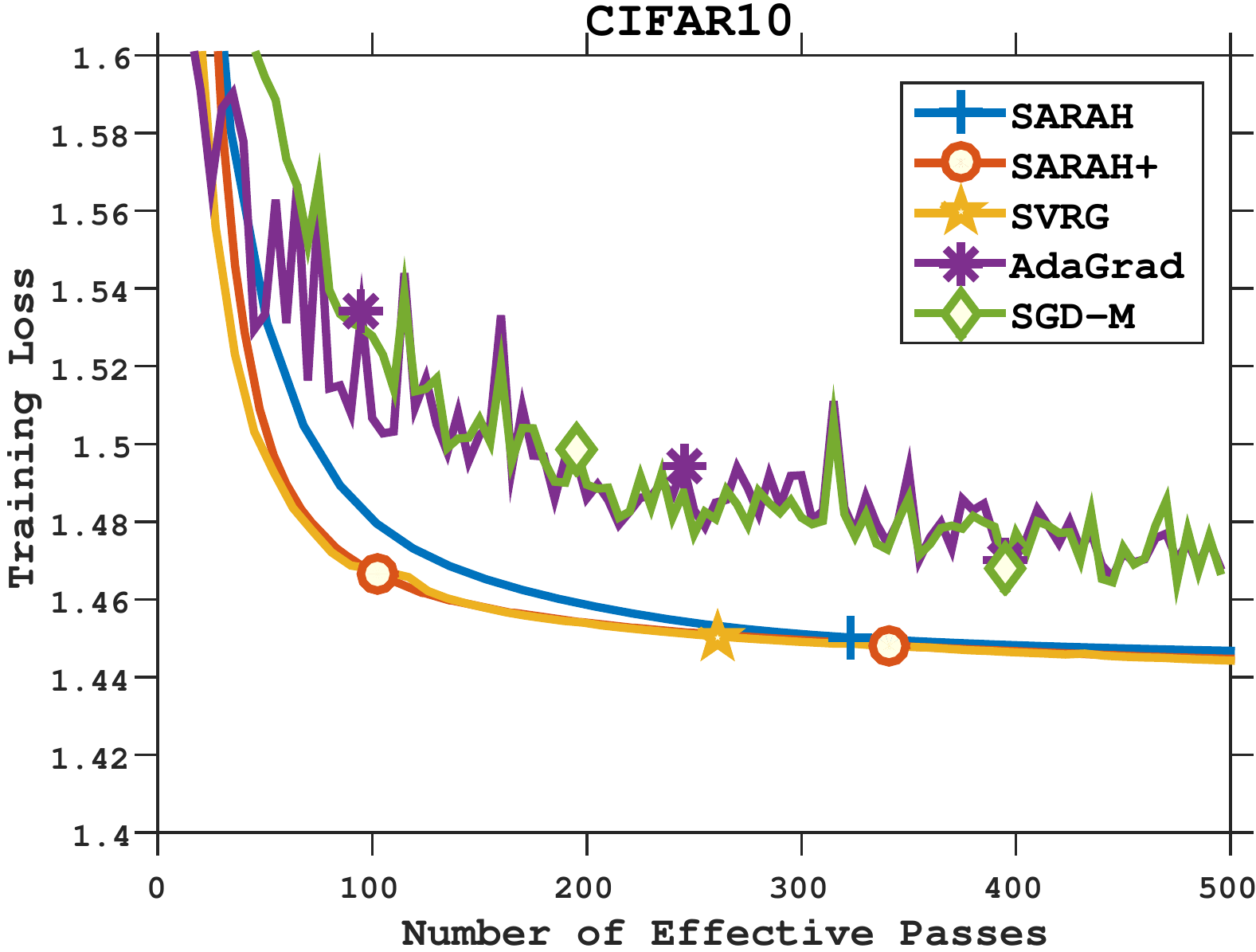,width=0.5\textwidth}
 \epsfig{file=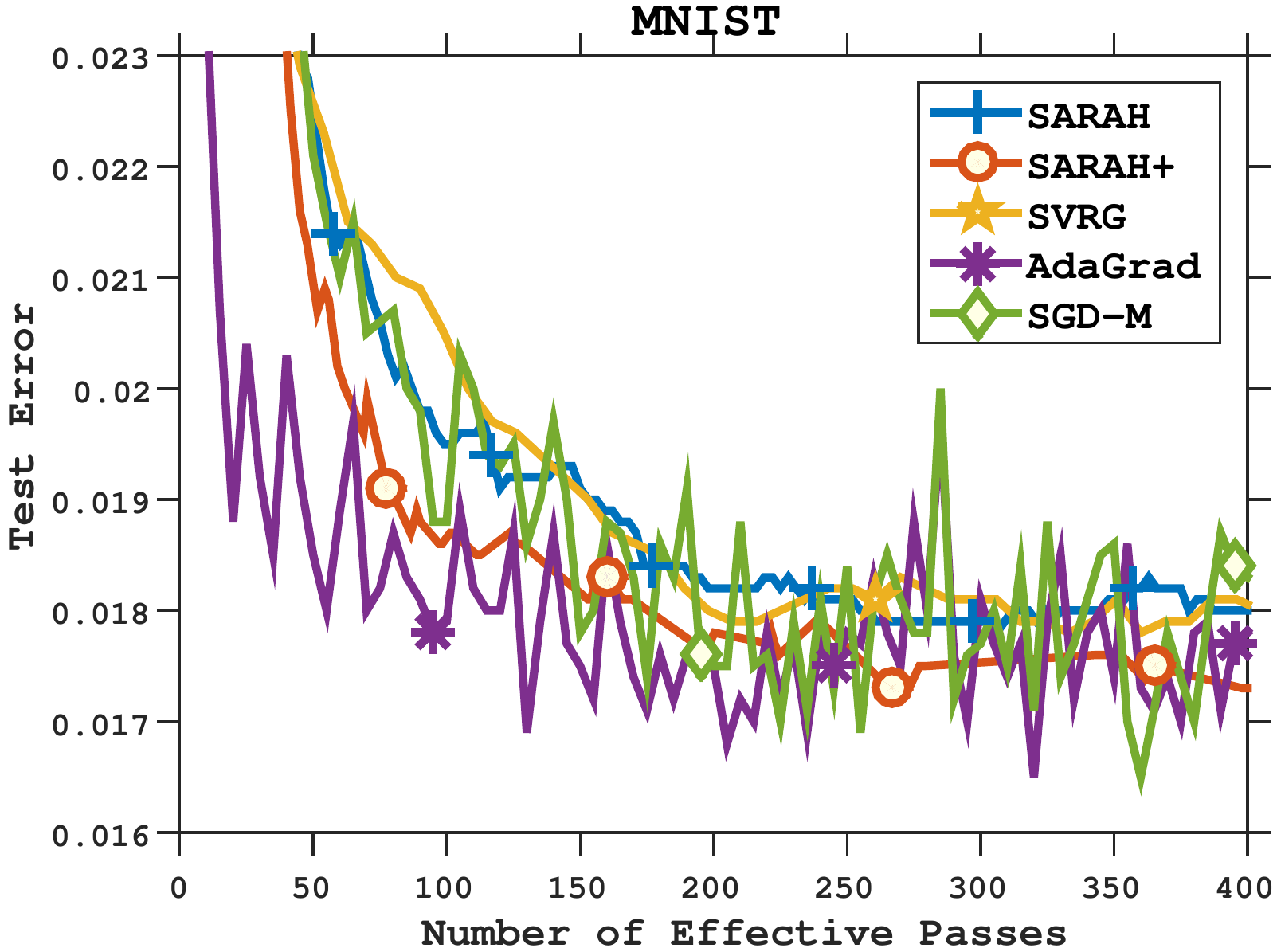,width=0.5\textwidth}
 \epsfig{file=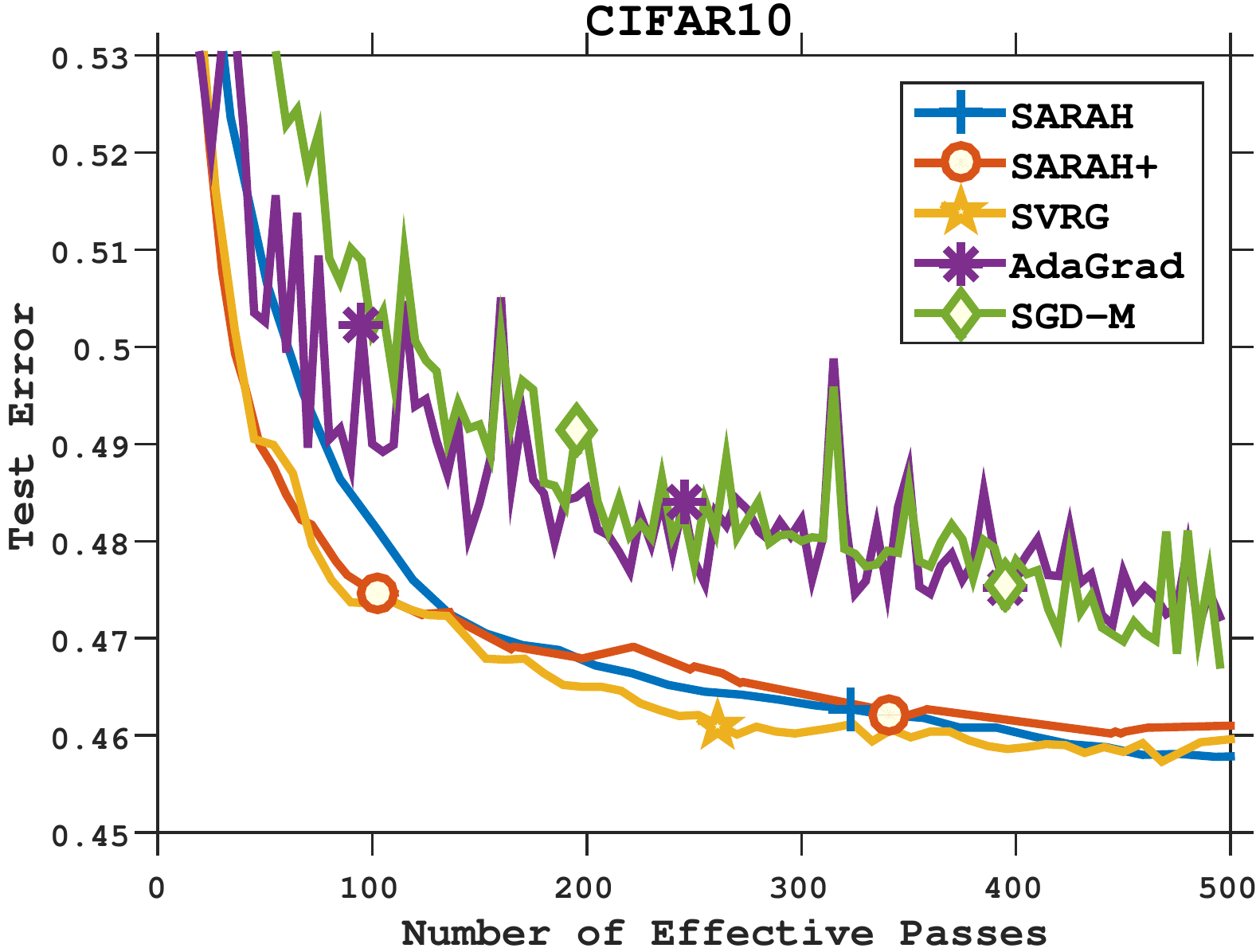,width=0.5\textwidth}
  \caption{\footnotesize An example of $\ell_2$-regularized neural nets on \emph{MNIST} and \emph{CIFAR10} training/testing datasets for SARAH, SARAH+, SVRG, AdaGrad and SGD-M.}
  \label{fig:exp1}
 \end{figure}

\paragraph{Performance and Comparison} We present the optimal choices of optimization parameters for the mentioned algorithms in Table~\ref{table:stats}, as well as their performance in Figure~\ref{fig:exp1}. As for the optimization parameters we consistently use the ratio $0.7$ in SARAH+, while for all the others, we need to tune two parameters, including $\eta^*$ for optimal learning rates, $m^*$ for optimal inner loop size, $\delta^*$ for the optimal initial accumulator and $\gamma^*$ for the optimal momentum. For the tuning of the parameters, reasonable ranges for the parameters have been scanned and we selected the best parameters in terms ofthe  training error reduction.

Figure~\ref{fig:exp1} compares the training losses (top) and the test errors (bottom), obtained by the tested algorithms on \emph{MNIST} and \emph{CIFAR10}, in terms of the number of effective passes through the data. On the   \emph{MNIST} dataset, which is deemed to be easier for traning, all the methods achieve similar performance in the end; however, SARAH(+) and SVRG stabilize faster than AdaGrad and SGD-M - the two of the most popular SGD variants; meanwhile, SARAH+ has shown superior performance in minimizing the training loss. For the other, more difficult, \emph{CIFAR10} dataset, SARAH(+) and SVRG improve upon the training accuracy considerably in comparison with AdaGrad and SGD-M, and as a result, a similar advantage can be seen in the test error reduction. 


\section{Conclusion} In this paper of work, we study and extend SARAH framework to nonconvex optimization,  also admitting the practical variant, SARAH+. For smooth nonconvex functions, the inner loop of SARAH achieves the best sublinear convergence rate in the literature, while the full variant of SARAH
has the same linear convergence rate and same as SVRG, for a special class of gradient dominated functions. In addition, we also analyze the dependence of the  convergence of SARAH on the size of the mini-batches. In the end, we validate SARAH(+) numerically in comparison with SVRG, AdaGrad and SGD-M, with the popular nonconvex application of neural networks.


\bibliography{reference}
\bibliographystyle{plainnat}

\appendix
\newpage


  \vbox{%
    \hsize\textwidth
    \linewidth\hsize
    \vskip 0.1in
  \hrule height 4pt
  \vskip 0.25in
  \vskip -5.5pt%
  \centering
    {\LARGE\bf{Supplementary Material} \par}
      \vskip 0.29in
  \vskip -5.5pt
  \hrule height 1pt
  \vskip 0.09in%
    
  \vskip 0.2in
    
  }

\section{Technical Proofs}

\subsection{Proof of Lemma \ref{lem_main_derivation_mb}}

By Assumption \ref{ass_Lsmooth} and $w_{t+1} = w_{t} - \eta v_{t}$, we have
\begin{align*}
\mathbb{E}[ P(w_{t+1})] & \overset{\eqref{eq:Lsmooth}}{\leq}  \mathbb{E}[ P(w_{t})] - \eta \mathbb{E}[\nabla P(w_{t})^\top v_{t}] 
+ \frac{L\eta^2}{2} \mathbb{E} [ \| v_{t} \|^2 ] 
\\
& = \mathbb{E}[ P(w_{t})] - \frac{\eta}{2} \mathbb{E}[ \| \nabla P(w_{t})\|^2 ] 
+ \frac{\eta}{2} \mathbb{E}[ \| \nabla P(w_{t}) - v_{t} \|^2 ] 
- \left( \frac{\eta}{2} - \frac{L\eta^2}{2} \right) \mathbb{E} [ \| v_{t} \|^2 ],
\end{align*}
where the last equality follows from the fact
$r^Tq = \frac{1}{2}\left[\|r\|^2 + \|q\|^2 - \|r-q\|^2\right],$ for any $r,q\in \R^d$.

By summing over $t = 0,\dots,m$, we have
\begin{align*}
\mathbb{E}[ P(w_{m + 1})] & \leq  \mathbb{E}[ P(w_{0})] - \frac{\eta}{2} \sum_{t=0}^{m} \mathbb{E}[ \| \nabla P(w_{t})\|^2 ] + \frac{\eta}{2} \sum_{t=0}^{m} \mathbb{E}[ \| \nabla P(w_{t}) - v_{t} \|^2 ]  
- \left( \frac{\eta}{2} - \frac{L\eta^2}{2} \right) \sum_{t=0}^{m} \mathbb{E} [ \| v_{t} \|^2 ],  
\end{align*}
which is equivalent to ($\eta>0$):
\begin{align*}
\sum_{t=0}^{m} \mathbb{E}[ \| \nabla P(w_{t})\|^2 ]  & \leq \frac{2}{\eta} \mathbb{E}[ P(w_{0}) - P(w_{m+1})] + \sum_{t=0}^{m} \mathbb{E}[ \| \nabla P(w_{t}) - v_{t} \|^2 ]  
 - ( 1 - L\eta ) \sum_{t=0}^{m} \mathbb{E} [ \| v_{t} \|^2 ] \\
& \leq \frac{2}{\eta} [ P(w_{0}) - P(w_{*})] + \sum_{t=0}^{m} \mathbb{E}[ \| \nabla P(w_{t}) - v_{t} \|^2 ]  
 - ( 1 - L\eta ) \sum_{t=0}^{m} \mathbb{E} [ \| v_{t} \|^2 ],    
\end{align*}

where the last inequality follows since $w_{*}$ is a global minimizer of $P$. (Note that $w_{0}$ is given.) 

\subsection{Proof of Lemma \ref{lem:var_diff_mb}}

Let $\mathcal{F}_{j} = \sigma(w_0,i_1,i_2,\dots,i_{j-1})$ be the $\sigma$-algebra generated by $w_0,i_1,i_2,\dots,i_{j-1}$; $\mathcal{F}_{0} = \mathcal{F}_{1} = \sigma(w_0)$. Note that $\mathcal{F}_{j}$ also contains all the information of $w_0,\dots,w_{j}$ as well as $v_0,\dots,v_{j-1}$. For $j \geq 1$, we have
\begin{align*}
\mathbb{E}[ \| \nabla P(w_{j}) - v_{j} \|^2 | \mathcal{F}_{j} ] 
& = \mathbb{E}[ \| [\nabla P(w_{j-1}) - v_{j-1} ] + [ \nabla P(w_{j}) - \nabla P(w_{j-1}) ]  - [ v_{j} - v_{j-1} ] \|^2 | \mathcal{F}_{j} ]
\\
& = \| \nabla P(w_{j-1}) - v_{j-1} \|^2 + \| \nabla P(w_{j}) - \nabla P(w_{j-1}) \|^2 + \mathbb{E} [ \| v_{j} - v_{j-1}  \|^2 | \mathcal{F}_{j} ] 
\\
&\quad + 2 ( \nabla P(w_{j-1}) - v_{j-1} )^\top ( \nabla P(w_{j}) - \nabla P(w_{j-1}) ) \\
&\quad - 2 ( \nabla P(w_{j-1}) - v_{j-1} )^\top \mathbb{E}[ v_{j} - v_{j-1} | \mathcal{F}_{j} ]  \\
&\quad - 2 ( \nabla P(w_{j}) - \nabla P(w_{j-1}) )^\top \mathbb{E}[ v_{j} - v_{j-1} | \mathcal{F}_{j} ] 
\\
& = \| \nabla P(w_{j-1}) - v_{j-1} \|^2 - \| \nabla P(w_{j}) - \nabla P(w_{j-1}) \|^2 + \mathbb{E} [ \| v_{j} - v_{j-1}  \|^2 | \mathcal{F}_{j} ], 
\end{align*}
where the last equality follows from
\begin{align*}
\mathbb{E}[ v_{j} - v_{j-1} | \mathcal{F}_{j} ] & \overset{\eqref{eq:vt_mb}}{= }\mathbb{E}\Big[ \frac{1}{b} \sum_{i \in I_{j}} [\nabla f_{i} (w_{j}) - \nabla f_{i}(w_{j-1})] \Big | \mathcal{F}_{j} \Big] \\
&= \frac{1}{b} \cdot \frac{b}{n} \sum_{i=1}^{n} [\nabla f_{i} (w_{j}) - \nabla f_{i}(w_{j-1})] = \nabla P(w_{j}) - \nabla P(w_{j-1}).
\end{align*}

By taking expectation for the above equation, we have
\begin{align*}
\mathbb{E}[ \| \nabla P(w_{j}) - v_{j} \|^2 ] &= \mathbb{E}[ \| \nabla P(w_{j-1}) - v_{j-1} \|^2 ] - \mathbb{E}[ \| \nabla P(w_{j}) - \nabla P(w_{j-1}) \|^2 ] + \mathbb{E}[ \| v_{j} - v_{j-1} \|^2 ]. 
\end{align*}

Note that $\| \nabla P(w_{0}) - v_{0} \|^2 = 0$. By summing over $j = 1,\dots,t\ (t\geq 1)$, we have
\begin{align*}
\mathbb{E}[ \| \nabla P(w_{t}) - v_{t} \|^2 ]  = \sum_{j = 1}^{t} \mathbb{E}[ \| v_{j} - v_{j-1} \|^2 ]  - \sum_{j = 1}^{t} \mathbb{E}[ \| \nabla P(w_{j}) - \nabla P(w_{j-1}) \|^2 ]. 
\end{align*}

\subsection{Proof of Lemma \ref{lem:var_diff_mb_02}}

Let
\begin{align*}
\xi_t = \nabla f_{t} (w_{j}) - \nabla f_{t}(w_{j-1}). \tagthis \label{xi_value}
\end{align*}

We have
\begin{align*}
& \mathbb{E}[\| v_j - v_{j-1} \|^2 | \mathcal{F}_{j} ] - \| \nabla P(w_{j}) - \nabla P(w_{j-1}) \|^2 \\
& \qquad \overset{\eqref{eq:vt_mb}}{=} \mathbb{E} \Big[ \Big\| \frac{1}{b} \sum_{i \in I_{j}} [\nabla f_{i} (w_{j}) - \nabla f_{i}(w_{j-1})] \Big\|^2 \Big| \mathcal{F}_{j} \Big] - \Big \| \frac{1}{n} \sum_{i=1}^{n} [\nabla f_{i} (w_{j}) - \nabla f_{i}(w_{j-1})] \Big \|^2 \\
& \qquad \overset{\eqref{xi_value}}{=} \mathbb{E} \Big[ \Big\| \frac{1}{b} \sum_{i \in I_{j}} \xi_i \Big\|^2 \Big| \mathcal{F}_{j} \Big] - \Big \| \frac{1}{n} \sum_{i=1}^{n} \xi_i \Big \|^2 \\
& \qquad = \frac{1}{b^2} \mathbb{E} \Big[ \sum_{i \in I_{j}} \sum_{k \in I_{j}} \xi_i^\top \xi_k  \Big| \mathcal{F}_{j} \Big] - \frac{1}{n^2} \sum_{i=1}^{n} \sum_{k=1}^{n} \xi_i^\top \xi_k \\
& \qquad = \frac{1}{b^2} \mathbb{E} \Big[ \sum_{i \neq k \in I_{j}} \xi_i^\top \xi_k + \sum_{i \in I_{j}} \xi_i^\top \xi_i  \Big| \mathcal{F}_{j} \Big] - \frac{1}{n^2} \sum_{i=1}^{n} \sum_{k=1}^{n} \xi_i^\top \xi_k \\
& \qquad = \frac{1}{b^2} \Big[ \frac{b}{n} \frac{(b-1)}{(n-1)} \sum_{i \neq k } \xi_i^\top \xi_k + \frac{b}{n}\sum_{i=1}^{n} \xi_i^\top \xi_i  \Big] - \frac{1}{n^2} \sum_{i=1}^{n} \sum_{k=1}^{n} \xi_i^\top \xi_k \\
& \qquad = \frac{1}{b^2} \Big[ \frac{b}{n} \frac{(b-1)}{(n-1)} \sum_{i=1}^{n} \sum_{k=1}^{n} \xi_i^\top \xi_k + \left( \frac{b}{n} - \frac{b}{n} \frac{(b-1)}{(n-1)} \right) \sum_{i=1}^{n} \xi_i^\top \xi_i  \Big] - \frac{1}{n^2} \sum_{i=1}^{n} \sum_{k=1}^{n} \xi_i^\top \xi_k \\
& \qquad = \frac{1}{b n} \Big[ \left( \frac{(b-1)}{(n-1)} - \frac{b}{n} \right) \sum_{i=1}^{n} \sum_{k=1}^{n} \xi_i^\top \xi_k + \frac{(n-b)}{(n-1)}  \sum_{i=1}^{n} \xi_i^\top \xi_i  \Big] \\
& \qquad = \frac{1}{b n} \left( \frac{n-b}{n-1} \right) \Big[ - \frac{1}{n}  \sum_{i=1}^{n} \sum_{k=1}^{n} \xi_i^\top \xi_k +  \sum_{i=1}^{n} \xi_i^\top \xi_i  \Big] \\
& \qquad = \frac{1}{b n} \left( \frac{n-b}{n-1} \right) \Big[ - n \Big\| \frac{1}{n}  \sum_{i=1}^{n} \xi_i \Big\|^2 +  \sum_{i=1}^{n} \| \xi_i \|^2  \Big] \\
& \qquad \leq \frac{1}{b} \left( \frac{n-b}{n-1} \right) \frac{1}{n}\sum_{i=1}^{n} \| \xi_i \|^2 \\
& \qquad \overset{\eqref{xi_value}}{=} \frac{1}{b} \left( \frac{n-b}{n-1} \right) \frac{1}{n}\sum_{i=1}^{n} \| \nabla f_{i} (w_{j}) - \nabla f_{i}(w_{j-1}) \|^2 \\
& \qquad \overset{\eqref{eq:Lsmooth_basic}}{\leq} \frac{1}{b} \left( \frac{n-b}{n-1} \right) L^2 \eta^2 \frac{1}{n} \sum_{i=1}^{n} \| v_{j-1} \|^2 \\
& \qquad = \frac{1}{b} \left( \frac{n-b}{n-1} \right) L^2 \eta^2 \| v_{j-1} \|^2. 
\end{align*}

Hence, by taking expectation, we have
\begin{align*}
\mathbb{E}[\| v_j - v_{j-1} \|^2 ] - \mathbb{E}[ \| \nabla P(w_{j}) - \nabla P(w_{j-1}) \|^2] \leq \frac{1}{b} \left( \frac{n-b}{n-1} \right) L^2 \eta^2 \mathbb{E}[ \| v_{j-1} \|^2]. 
\end{align*}

By Lemma \ref{lem:var_diff_mb}, for $t \geq 1$, 
\begin{align*}
\mathbb{E}[ \| \nabla P(w_{t}) - v_{t} \|^2 ] 
& = \sum_{j = 1}^{t} \mathbb{E}[ \| v_{j} - v_{j-1} \|^2 ]  
 - \sum_{j = 1}^{t} \mathbb{E}[ \| \nabla P(w_{j}) - \nabla P(w_{j-1}) \|^2 ] \\
 & \leq \frac{1}{b} \left( \frac{n-b}{n-1} \right) L^2 \eta^2 \sum_{j = 1}^{t} \mathbb{E}[ \| v_{j-1} \|^2].  
\end{align*}

This completes the proof. 

However, the result simply follows for the case when $b=1$ by the alternative proof. We have
\begin{align*}
\| v_{t} - v_{t-1} \|^2 &\overset{\eqref{eq:vt}}{=} \| \nabla f_{i_{t}} (w_{t}) - \nabla f_{i_{t}}(w_{t-1}) \|^2 
\overset{\eqref{eq:Lsmooth_basic}}{\leq} L^2 \| w_{t} - w_{t-1} \|^2 = L^2 \eta^2 \| v_{t-1} \|^2, \ t \geq 1. \tagthis \label{eq:afsag242}
\end{align*}

Hence, by Lemma \ref{lem:var_diff_mb}, we have
\begin{align*}
\mathbb{E}[ \| \nabla P(w_{t}) - v_{t} \|^2 ] & \leq \sum_{j = 1}^{t} \mathbb{E}[ \| v_{j} - v_{j-1} \|^2 ] \overset{\eqref{eq:afsag242}}{\leq} L^2 \eta^2 \sum_{j = 1}^{t} \mathbb{E}[ \| v_{j - 1} \|^2 ]. 
\end{align*}

\subsection{Proof of Theorem \ref{thm:nonconvex_01_mb}}

By Lemma \ref{lem:var_diff_mb_02}, we have
\begin{align*}
\mathbb{E}[ \| \nabla P(w_{t}) - v_{t} \|^2 ]  \leq \frac{1}{b} \left( \frac{n-b}{n-1} \right) L^2 \eta^2 \sum_{j=1}^{t} \mathbb{E}[\| v_{j-1} \|^2]. 
\end{align*}

Note that $\| \nabla P(w_{0}) - v_{0} \|^2 = 0$. Hence, by summing over $t = 0,\dots,m$ ($m \geq 1$), we have
\begin{align*}
\sum_{t=0}^{m} \mathbb{E}\| v_{t} - \nabla P(w_{t}) \|^2 \leq \frac{1}{b} \left( \frac{n-b}{n-1} \right) L^2 \eta^2 \Big[ m  \mathbb{E}\|v_{0} \|^2 + (m-1) \mathbb{E}\|v_{1} \|^2 + \dots + \mathbb{E}\|v_{m-1}\|^2 \Big ]. 
\end{align*}

We have
\begin{align*}
& \sum_{t=0}^{m} \mathbb{E}[ \| \nabla P(w_{t}) - v_{t} \|^2 ]  - ( 1 - L\eta ) \sum_{t=0}^{m} \mathbb{E} [ \| v_{t} \|^2 ] \\
& \leq \frac{1}{b} \left( \frac{n-b}{n-1} \right) L^2 \eta^2 \Big[ m  \mathbb{E}\|v_{0} \|^2 + (m-1) \mathbb{E}\|v_{1} \|^2 + \dots + \mathbb{E}\|v_{m-1}\|^2 \Big ] \\ &- (1 - L\eta) \Big [ \mathbb{E}\|v_{0} \|^2 + \mathbb{E}\|v_{1} \|^2 + \dots + \mathbb{E}\|v_{m}\|^2  \Big ] \\
& \leq \Big[\frac{1}{b} \left( \frac{n-b}{n-1} \right) L^2\eta^2 m - (1 - L\eta) \Big] \sum_{t=1}^{m} \mathbb{E} [ \| v_{t-1} \|^2 ] \overset{\eqref{eta_mb}}{\leq} 0,  \tagthis \label{eq:equal_zero_mb}
\end{align*}

since 
\begin{align*}
\eta = \frac{2}{L\left(\sqrt{1 + \frac{4m}{b}\left(\frac{n-b}{n-1} \right)} + 1\right)}
\end{align*}

is a root of equation 
\begin{align*}
\frac{1}{b} \left( \frac{n-b}{n-1} \right) L^2\eta^2 m - (1 - L\eta) = 0. 
\end{align*}

Therefore, by Lemma \ref{lem_main_derivation_mb}, we have
\begin{align*}
\sum_{t=0}^{m} \mathbb{E}[ \| \nabla P(w_{t})\|^2 ] & \leq \frac{2}{\eta} [ P(w_{0}) - P(w_{*})] + \sum_{t=0}^{m} \mathbb{E}[ \| \nabla P(w_{t}) - v_{t} \|^2 ]  
 - ( 1 - L\eta ) \sum_{t=0}^{m} \mathbb{E} [ \| v_{t} \|^2 ] \\
 & \overset{\eqref{eq:equal_zero_mb}}{\leq} \frac{2}{\eta} [ P(w_{0}) - P(w_{*})].  
\end{align*}

If $\tilde{w} = w_t$, where $t$ is chosen uniformly at random from $\{0,1,\dots,m\}$, then 
\begin{align*}
\mathbb{E}[ \| \nabla P(\tilde{w})\|^2 ] = \frac{1}{m+1}\sum_{t=0}^{m} \mathbb{E}[ \| \nabla P(w_{t})\|^2 ] \leq \frac{2}{\eta(m+1)} [ P(w_{0}) - P(w_{*})].  
\end{align*}

\subsection{Proof of Theorem \ref{thm:nonconvex_02_mb}}

Note that $\tilde{w}_s = \tilde{w}$ and $w_{0} = \tilde{w}_{s-1}$, $s \geq 1$. By Theorem \ref{thm:nonconvex_01_mb}, we have
\begin{align*}
\mathbb{E}[ \| \nabla P(\tilde{w}_s)\|^2 | \tilde{w}_{s-1} ] = \mathbb{E}[ \| \nabla P(\tilde{w})\|^2 | w_{0} ] & \leq \frac{2}{\eta(m+1)} [ P(w_{0}) - P(w_{*})] \\
& \overset{\eqref{gradientdominated}}{\leq} \frac{2\tau}{\eta(m+1)}  \| \nabla P(w_{0}) \|^2  \\
& = \frac{2\tau}{\eta(m+1)}  \| \nabla P(\tilde{w}_{s-1}) \|^2 . 
\end{align*}

Hence, taking expectation to have
\begin{align*}
\mathbb{E}[ \| \nabla P(\tilde{w}_s)\|^2 ] \leq \frac{2\tau}{\eta(m+1)} \mathbb{E}[ \| \nabla P(\tilde{w}_{s-1}) \|^2 ] \leq \left[ \frac{2\tau}{\eta(m+1)}  \right]^s \| \nabla P(\tilde{w}_{0}) \|^2.
\end{align*}

\section{Additional Plots}
\paragraph{Performance on \emph{MNIST} with $b=20$} In addition to the plots in Section~\ref{sec:exp} with mini-batch size $b=10$, we also experimented with $b=20$ for \emph{MNIST} where the other settings of the network remain the same. Similar trend can be observed for the algorithms, where SARAH+ seems to be the best in terms of sub-optimality/training loss while SVRG and SARAH follows with roughly comparable performance. The two SGD variants--AdaGrad and SGD-M exhibit a little worse performance.

\begin{figure}[H] 
 \epsfig{file=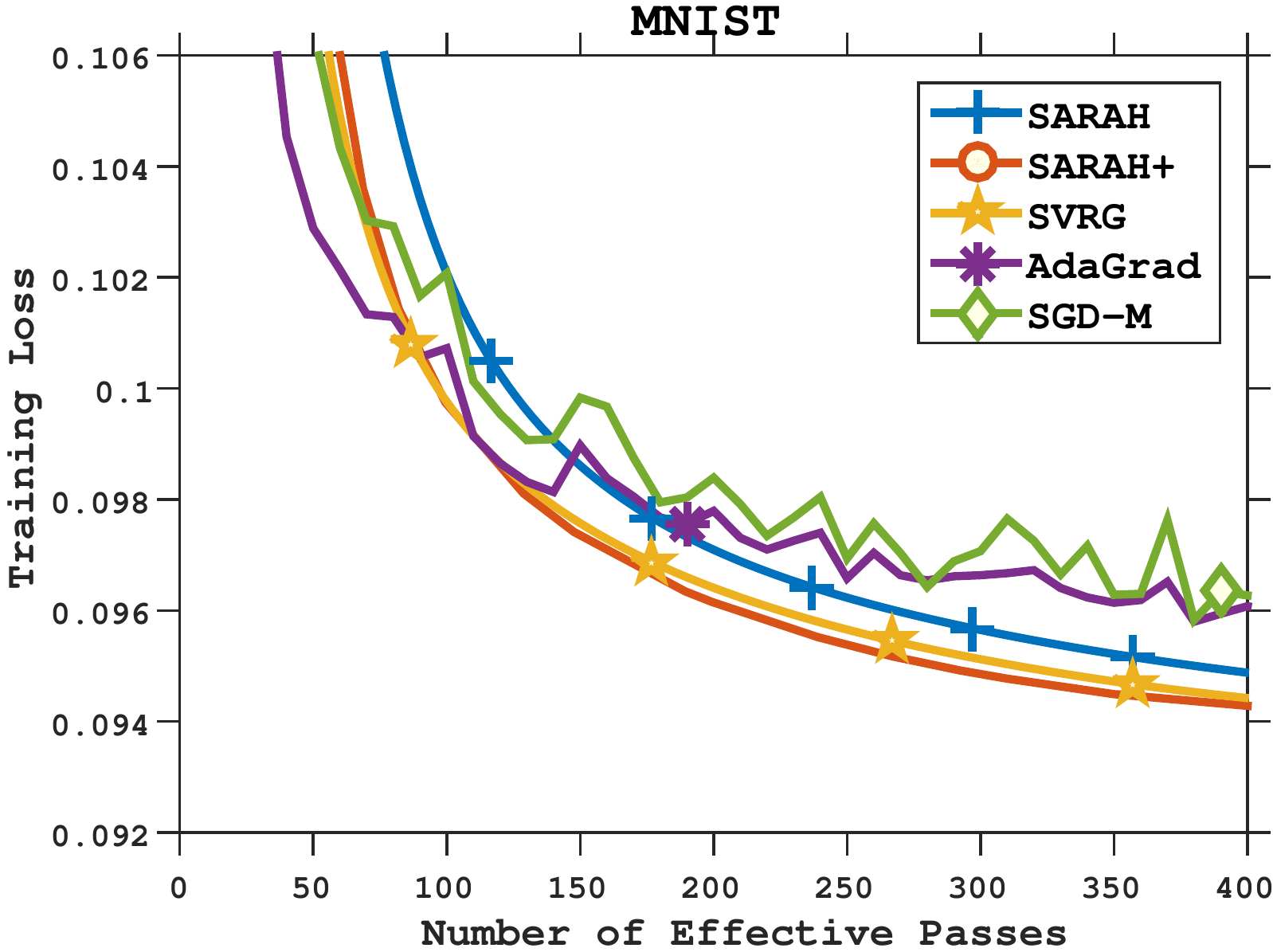,width=0.5\textwidth}
 \epsfig{file=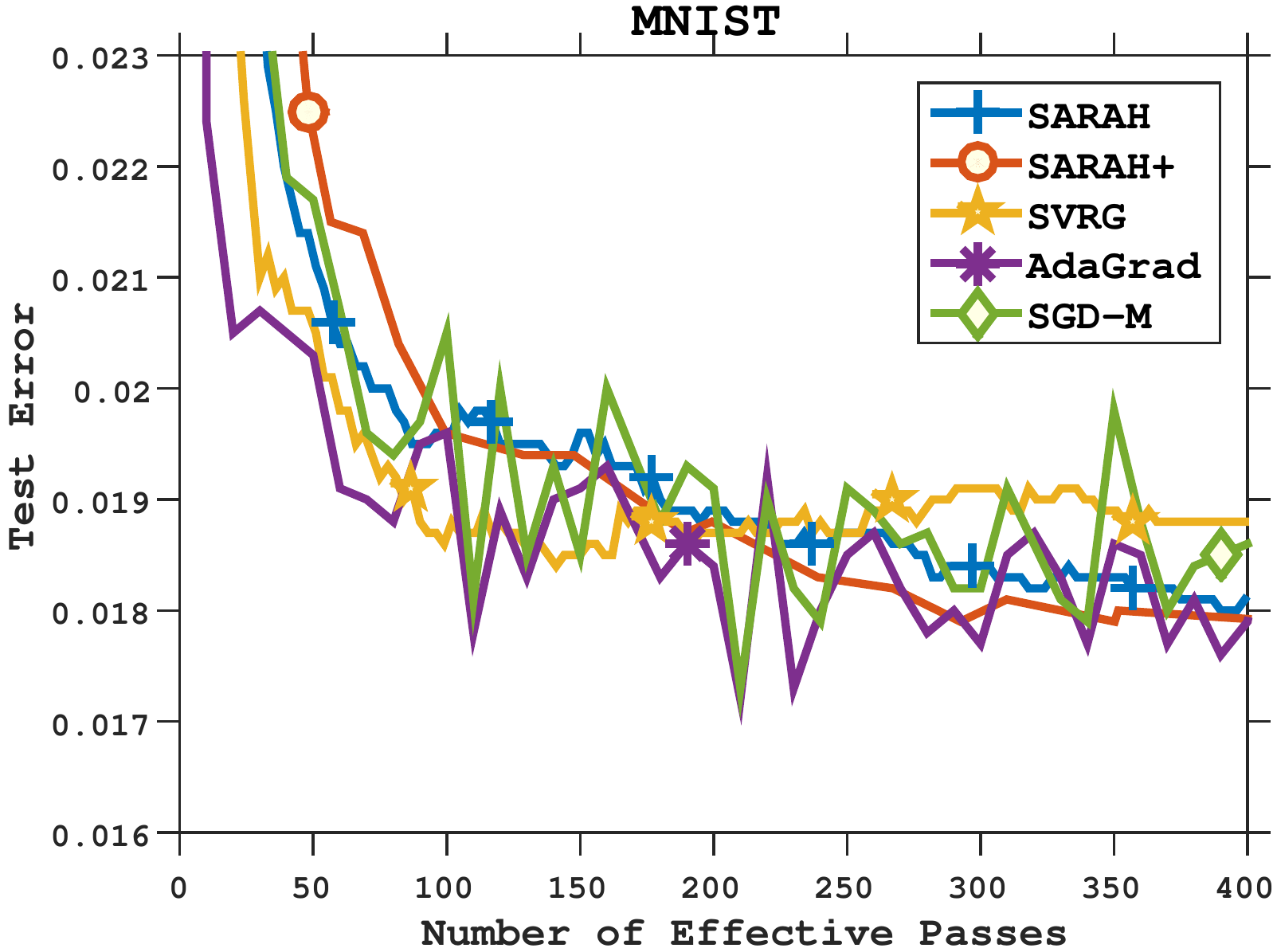,width=0.5\textwidth}
  \caption{\footnotesize An example of $\ell_2$-regularized neural nets on \emph{MNIST} training dataset for SARAH, SARAH+, SVRG, AdaGrad and SGD-M with mini-batch size $b=20$.}
  \label{fig:exp2}
 \end{figure}
 
\begin{figure}[H] 
 \epsfig{file=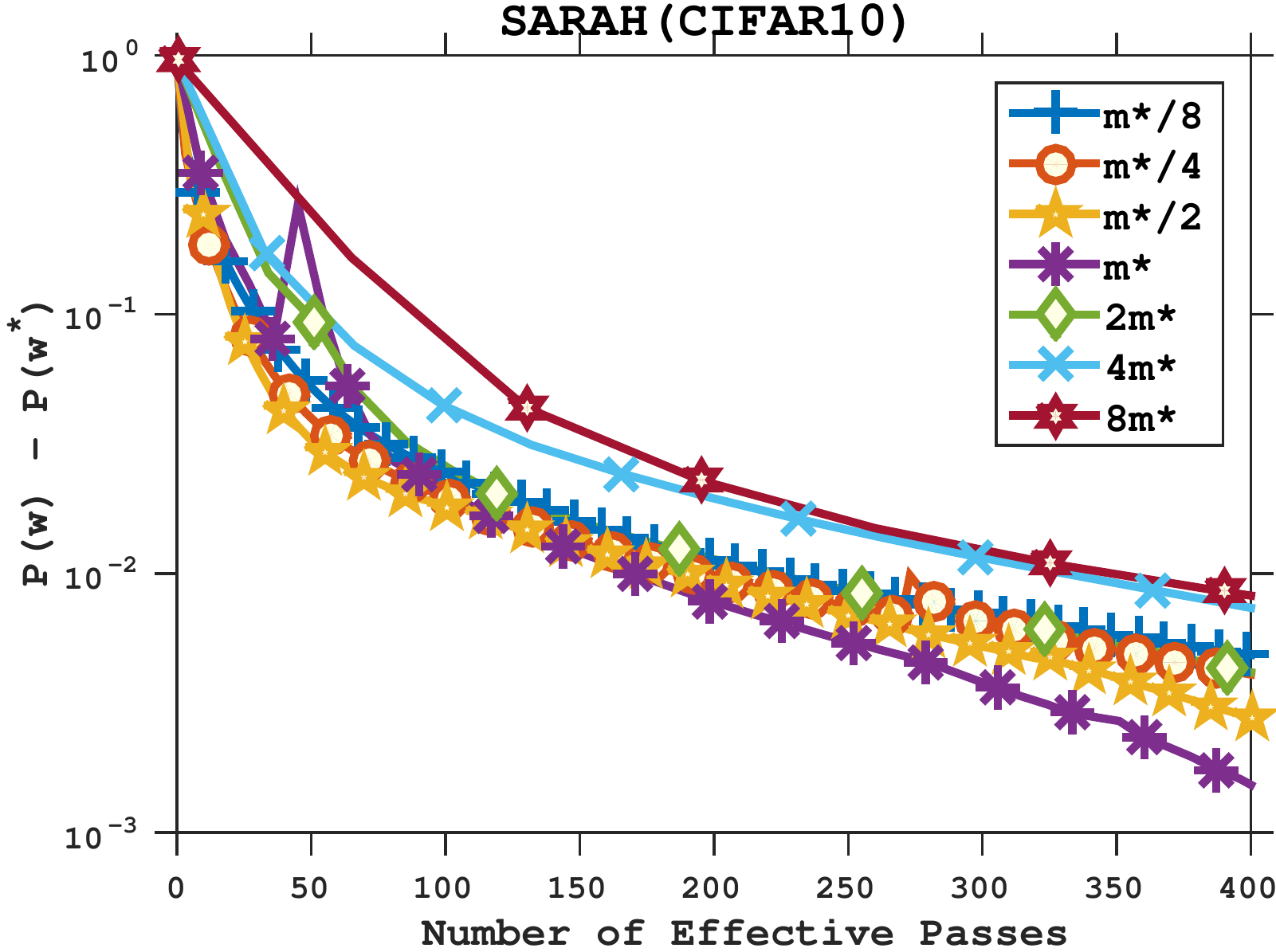,width=0.5\textwidth}
 \epsfig{file=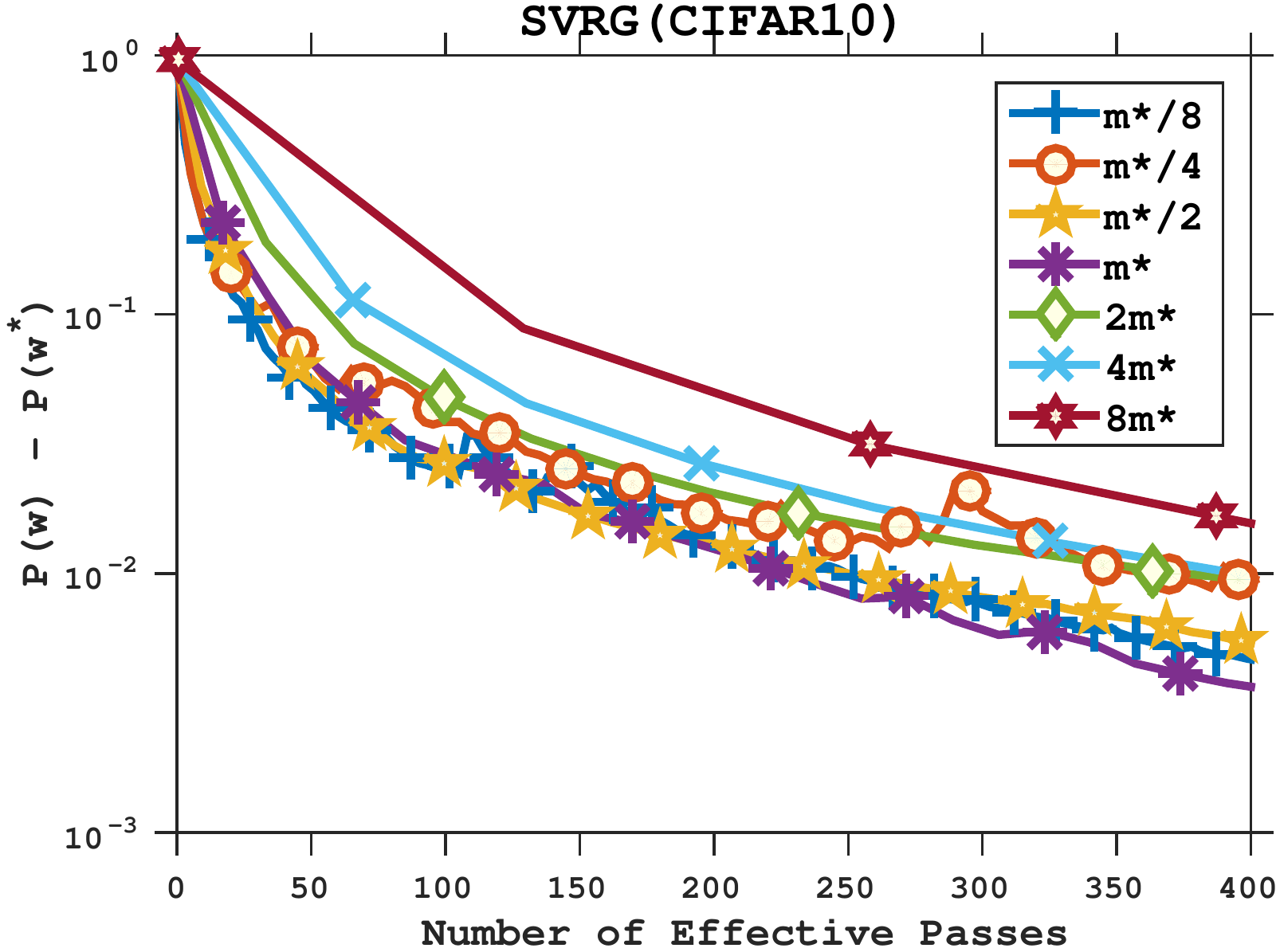,width=0.5\textwidth}
  \caption{\footnotesize Sensitivity analysis of the inner loop size $m$ for SARAH and SVRG with the example of $\ell_2$-regularized neural nets on \emph{CIFAR10} training dataset.}
  \label{fig:exp3}
 \end{figure}

 \paragraph{Sensitivity of Inner Loop Size on \emph{CIFAR10}} To validate the necessity of SARAH+, we performance sensitivity analyses for the inner loop size $m$, that is we show the importance of the choice of this $m$ for SARAH and SVRG on \emph{CIFAR10} in Figure~\ref{fig:exp3}, where $m^*$s denote the best choices what are presented in Table~\ref{table:stats}.

%
%
%
%

\end{document}